\documentclass[review]{elsarticle}

\usepackage{lineno,hyperref}
\modulolinenumbers[5]

\usepackage{times}
\usepackage{epsfig}
\usepackage{graphicx}
\usepackage{amsmath}
\usepackage{amssymb}
\usepackage{color,soul}
\graphicspath{fig/}
\usepackage{subcaption}
\usepackage{booktabs}

\journal{Journal of Pattern Recognition}

\begin{document}

\begin{frontmatter}

\title{Certainty Driven Consistency Loss on Multi-Teacher Networks for Semi-Supervised Learning}

\author[mymainaddress]{Lu Liu\corref{mycorrespondingauthor}}
\cortext[mycorrespondingauthor]{Corresponding author}
\ead{lliu@u.nus.edu}

\author[mymainaddress,mysecondaryaddress]{Robby T. Tan}
\ead{robby.tan@yale-nus.edu.sg}

\address[mymainaddress]{Electrical and Computer Engineering Department, National University of Singapore\\
E4-06-20, 4 Engineering Drive 3, 117583, Singapore}
\address[mysecondaryaddress]{Yale-NUS College, Singapore}

\begin{abstract}
One of the successful approaches in semi-supervised learning is based on the consistency regularization. Typically, a student model is trained to be consistent with teacher prediction for the inputs under different perturbations. To be successful, the prediction targets given by teacher should have good quality, otherwise the student can be misled by teacher. Unfortunately, existing methods do not assess the quality of the teacher targets. In this paper, we propose a novel Certainty-driven Consistency Loss (CCL) that exploits the predictive uncertainty in the consistency loss to let the student dynamically learn from reliable targets.
Specifically, we propose two approaches, i.e. Filtering CCL and Temperature CCL to either filter out uncertain predictions or pay less attention on them in the consistency regularization. 
We further introduce a novel decoupled framework to encourage model difference. 
Experimental results on SVHN, CIFAR-10, and CIFAR-100 demonstrate the advantages of our method over a few existing methods.
\end{abstract}

\begin{keyword}
semi-supervised learning \sep certainty-driven consistency loss \sep uncertainty estimation \sep decoupled student-teacher \sep reliable targets \sep noisy labels
\end{keyword}

\end{frontmatter}


\section{Introduction}
Deep neural networks achieve tremendous success in many visual tasks such as image recognition~\cite{he2016deep} and object detection~\cite{ren2015faster}. However, training networks requires large-scale labeled datasets~\cite{russakovsky2015imagenet,lin2014microsoft} which are usually expensive and difficult to collect. Semi-supervised learning (SSL) aims to boost the model performance by leveraging a limited amount of labeled data and a large amount of unlabeled data ~\cite{chapelle2009semi,zhu2006semi}.

Most of the recent methods follow some noise regularization approaches, e.g. data augmentation and dropout, to encourage the model to give similar predictions under random perturbations. The network is trained using a standard supervised classification loss and an unsupervised consistency loss. Pseudo ensembles~\cite{bachman2014learning} and $\Gamma$-model in the ladder network~\cite{rasmus2015semi} produce a noisy student model and clean teacher model, and train the student to predict consistently with the "soft targets" (i.e. softmax probability distributions) generated by the teacher. Following this student-teacher framework, $\Pi$ model~\cite{laine2016temporal} simplifies the network by making both student and teacher model noisy. Recent works focus on designing a better teacher model. Temporal ensemble~\cite{laine2016temporal} and mean teacher (MT)~\cite{tarvainen2017mean} generate a better teacher by prediction ensemble and model weights ensemble, respectively. FSWA~\cite{athiwaratkun2019there} proposes fast stochastic weight averaging to obtain a stronger ensemble teacher faster.

One common issue of the perturbation-based SSL approach is a problem called confirmation bias~\cite{tarvainen2017mean}, which means the model is prone to confirm the previous predictions and resist new changes. If the unsupervised consistency loss outweighs the supervised classification loss, the model cannot learn any meaningful knowledge, and hence can get stuck in a degenerated solution. Most existing methods~\cite{laine2016temporal,miyato2018virtual,tarvainen2017mean,luo2018smooth} address this issue by employing a weighting function that gives a ramp-up weight (i.e. a gradually increasing weight) for the consistency loss. This ramp-up loss weight can let the model learn more from supervised loss early in the training, and then gradually learn from the unsupervised consistency regularization.

However, we consider that solely using a ramp-up weight is still ineffective to solve the confirmation bias. Even though the ramp-up weight is used, the unsupervised consistency loss is applied to all training samples blindly, ignoring the fact that not all training data provide meaningful and reliable information. In the context of the student-teacher framework, it means the student blindly learns from all noisy targets, regardless of the quality of the teacher's targets. For labeled data, the supervised classification loss probably can correct some of the mistakes in the successive training steps. However, for the majority of unlabeled data, they can still remain in the previously enforced wrong predictions, since no ground-truth class labels are available to correct their predictions.

In this paper, our goal is to let the student gradually learn from meaningful and reliable targets from the teacher, rather than some noisy misleading information. We propose a Certainty-driven Consistency Loss (CCL) to exploit uncertainty information when enforcing the consistency between perturbated predictions given by the student and the teacher. In general, there are two ways to tackle uncertain targets: hard filtering, and soft weighting/attention. Our basic idea is that if the teacher is uncertain about its prediction of a training sample, the teacher should either filters it out from the student's learning list, or let the student learn with a lesser effort from it. We present two approaches to utilize CCL: \textit{Filtering CCL} and \textit{Temperature CCL}, to enforce consistency in either a hard or a soft way.

Principally, our method lets the student dynamically learn from more meaningful and reliable targets. Our method adapts a progressive learning strategy~\cite{bengio2009curriculum} to gradually learn from certain targets to less certain targets in the consistency regularization. Unlike existing SSL methods that blindly enforce consistency loss on all unlabeled data, our method enforces consistency gradually from high certainty regions to low certainty regions through our hard filtering and soft temperature consistency regularization.

Our method is also in conformity to the smoothness assumption~\cite{chapelle2009semi}, that is: "If two points $x_1, x_2$ in a high-density region are close, then so should be the corresponding outputs $y_1, y_2$". This assumption implies that points in a high-density region should generate more consistent predictions than those in a low-density region. 
Inspired by this implication, our key idea is to enhance stronger consistency regularization in the high-density regions than in the low-density regions. 
We consider that it is better to enhance the consistency loss gradually from high-density regions to low-density regions, with higher learning weights in high-density regions, rather than enhancing the consistency in all regions equally. To achieve this, our method first estimates the predictive certainty of each data, which has positive correlation with the data local density. The reason is that if a data point lies in a high-density region (e.g. cluster center), the predictive certainty should be high. On the other hand, if a data point lies in a low-density region (e.g. cluster boundary), the predictive certainty should be low. 
Then, based on the certainty estimation, we propose two approaches in both hard and soft ways to achieve the attentive consistency regularization.
Specifically, our Filtering CCL adopts a curriculum strategy, where we gradually learn from certain unlabeled predictions to less certain ones. Our Temperature CCL scales down the loss magnitude of uncertain predictions, and hence the erroneous gradients coming from the unreliable unlabeled data can be reduced.

To let the teacher provide additional useful information for the student, we further propose a decoupled consistency in a multi-teacher framework, where the consistency regularization is enforced between decoupled students and teachers with different network initialization and training conditions. We argue that the strongly coupled student-teacher in the existing perturbation-based methods~\cite{laine2016temporal,tarvainen2017mean} can limit the capacity of the model, due to the high similarity between them. We introduce a framework to decouple the students and teachers, by forming them in a closed circle. In this way, the teacher does not teach the student that generated it anymore, but teaches the next student in a circle manner.

Our contribution is four-fold: 
(1) We present a certainty-driven consistency loss (CCL) that exploits the uncertainty of the model predictions for the consistency regularization, which has not been explored before in semi-supervised learning.   
(2) We propose two novel approaches Filtering CCL and Temperature CCL: Filtering CCL enforces consistency on the reliable targets by filtering out uncertain predictions; Temperature CCL reduces the magnitudes of gradients on the uncertain targets and lets the model pay more attention on learning certain ones.
(3) We introduce a combined method, FT-CCL, that utilizes both approaches and shows robustness to noisy labels. Since the noisy labels can be ignored in both hard and soft way, and thus the model can benefit from the most reliable predictions. Extensive experiments demonstrate the effectiveness of our proposed method. 
(4) We introduce a decoupled multi-teacher framework to encourage the teacher provide additional new knowledge for the student. The decoupled consistency and multi-teacher framework further boosts the performance.
\section{Related work} 

\paragraph{Semi-supervised learning}
Self-training is one of the earliest methods which incrementally adds unlabeled data with self-predicted labels with high confidence~\cite{scudder1965probability, rosenberg2005semi}. Co-training~\cite{blum1998combining,nigam2000analyzing} trains multi-view models for the classification task by using disjoint splits or views of the training data or other advanced techniques (e.g. image region division~\cite{HONG201559}). Due to the efficiency and simpleness, co-training framework has been used in many application tasks besides image classification, such as image segmentation~\cite{PENG2020107269,luo2019taking}, detection~\cite{SU2020107003}, tracking~\cite{ZHANG201782}, and knowledge distillation~\cite{ZHANG2021107659}. Recently, Qiao et al. propose deep Co-Training~\cite{qiao2018deep} which adds a consistency constraint on the adversarial examples between multi-view independent models. Both of our decoupled multi-teacher framework and deep Co-Training~\cite{qiao2018deep} follow the general co-training architecture~\cite{blum1998combining} that trains multiple independent models in parallel. However, the goals and mechanisms used to encourage model diversity are different. Deep co-training~\cite{qiao2018deep} encourages model difference among multiple student models using adversarial examples.
In our decoupled multi-teacher framework, our goal is to encourage the difference between the student and teacher. We decouple the student from the self-generated exponential moving average (EMA) teacher by letting it learn from another teacher, which is generated by the next student in the circle. Meanwhile, different student models are trained with different network initialization, random perturbations, and network dropout. 
Hence, the multiple students can provide different information about each data.
One advantage of our decoupled student-teacher framework over~\cite{qiao2018deep} and existing co-training methods is that, for each view, the single model performance is increased by the ``shadow" EMA teacher model with zero learning costs.

Several recently proposed methods are based on training the model predictions to be consistent to perturbations~\cite{miyato2018virtual,tarvainen2017mean,luo2018smooth,athiwaratkun2019there}. Pseudo ensembles~\cite{bachman2014learning} and $\Gamma$-model in the ladder network~\cite{rasmus2015semi} produce a noisy student and clean teacher, and trains the student model to predict the target given by the teacher. Following the same paradigm, $\Pi$ model~\cite{laine2016temporal} applies noise to both the student and the teacher, then penalizes inconsistent predictions. Temporal ensemble~\cite{laine2016temporal} penalizes the inconsistency between the network predictions and the temporally ensembled network predictions, and maintains an exponential moving average (EMA) prediction for each training data. Mean teacher~\cite{tarvainen2017mean} utilizes EMA on the model weights to maintain an averaged teacher model for generating targets for the student to learn from. VAT~\cite{miyato2018virtual} utilizes virtual adversarial training method to select the perturbations in the direction sensitive to the prediction of the classifier. Athiwaratkun et al.~\cite{athiwaratkun2019there} modify the stochastic weight averaging (SWA)~\cite{izmailov2018averaging} to obtain a stronger ensemble teacher faster. SEGCN~\cite{luo2020every} combines graph convolutional network (GCN) with mean teacher~\cite{tarvainen2017mean}, and builds a student graph and a teacher graph on top of the student and teacher model, respectively to model graph-structured data. Deep label propagation~\cite{2019dlp} combines mean teacher with transductive label propagation to infer pseudo labels for unlabeled data. However, these methods do not assess the quality of the targets, which can be unreliable for the student to learn from. In contrast, our approach leverages the uncertainty of perturbated predictions to provide reliable consistency constraints.

The recently proposed Fixmatch~\cite{sohn2020fixmatch} and DS~\cite{ke2019dual} share some merits with our method in terms of selecting the reliable or stable pseudo labels for the unlabeled data. However, our approach has advantage over these methods in the respect of selection indicator, and the attention mechanism among unlabeled data. 
Firstly, both Fixmatch and DS use the confidence score of the argmax class as the selection indicator. However, it has been shown that the deep neural network tends to be over-confident~\cite{gal2016dropout,guo2017calibration} (\textit{i.e.} a model can be uncertain in its predictions even with a high softmax output), making the predicted confidence score non-linear to the test accuracy. In contrast, our method uses Bayesian uncertainty, specifically Monte Carlo dropout~\cite{gal2016dropout,gal2015bayesian} to estimate the uncertainty of the unlabeled data predictions. We propose four uncertainty metrics, e.g. predictive variance, over multiple predictions under random input perturbations and network dropout, and hence, our method measures both data uncertainty and model uncertainty~\cite{kendall2017uncertainties}. The experimental results show that our proposed uncertainty estimation metrics are proportional to the error rate, and hence can be used as an error estimate.
Furthermore, both Fixmatch and DS rely on hard thresholding to filter out unreliable and unstable pseudo labels, without considering the relative uncertainty degree among them. In contrast, besides hard filtering,  in our proposed Temperature CCL, we use soft attention to control the loss contributions of different unlabeled data dynamically during the training phase according to their respective uncertainty levels.

\paragraph{Uncertainty modelling}
A few uncertainty modelling methods have been proposed based on Bayesian neural network~\cite{li2015stochastic,gal2016dropout,lakshminarayanan2017simple,springenberg2016bayesian}. One of them is Monte Carlo dropout proposed by~\cite{gal2016dropout,gal2015bayesian}. They theoretically prove that dropout at test time can be used to approximate a model's uncertainty. However, this uncertainty is only used during testing, which does not influence the training process. We extend this uncertainty modelling approach into training to evaluate the uncertainty of the predictions in SSL. Furthermore, our uncertainty also takes the local smoothness into consideration, in order to model both data uncertainty and model uncertainty~\cite{kendall2017uncertainties}.

\paragraph{Curriculum learning and active learning} 
Our method adapts progressive learning strategy and an uncertainty indicator to let the student gradually learn from certain targets to less certain ones, which relate to curriculum learning~\cite{bengio2009curriculum} and active learning ~\cite{settles2009active}. Bengio et al.~\cite{bengio2009curriculum} first propose a progressive learning paradigm which organizes the training data from easiest to hardest. In active learning~\cite{settles2009active,matiz2019inductive}, similar indicators have been used to guide decisions about which data point to label next. In general, we all rely on some indicators to evaluate the training data. However, to our knowledge, applying this indicator in terms of the consistency regularization to SSL is new, and has not been explored before.

\paragraph{Knowledge distillation}  
Hinton et al.~\cite{hinton2015distilling} apply the concept of temperature in model distillation, which aims to distill the knowledge from a large pre-trained network to a much smaller network without lossing much of the generalization ability. The temperature, a hyperparameter inside softmax function, is used to soften the probability distributions of softmax, which encourages the small model to learn more "dark knowledge" distributions from the large model, rather than the hard label. However, the method needs to set the value of temperature empirically, which is shared by all training samples. Our method can automatically define the temperature of each training sample according to its uncertainty, and use its own temperature to decide how much influence it has on training the student model.

\section{Our Approach}

One common drawback of the existing perturbation-based methods is that they regularize the outputs to be smooth regardless of the quality of the targets. We address this drawback by estimating the uncertainty of the targets, and then let the student learn more from certain targets, and less from uncertain targets. We achieve this by either filtering out uncertain targets, or decreasing the relative impact of uncertain targets vs. certain ones. By doing this, the student learns meaningful and reliable knowledge instead of some error prone information. Our certainty driven consistency loss improves the student model, which in turn forms a better teacher model that can generate high-quality targets. 

Let the whole training set $D$ consist of total of $N$ examples, out of which only $N_l$ are labeled. Let $D_L=\{(x_i,y_i)\}_{i=1}^{N_l}$ be the labeled set, where $y_i \in \{1,...,C\}$, $C$ is the number of classes, and $D_U$ be the unlabeled set. Given an input batch $B$ with $|B|$ training samples, we aim to minimize the supervised classification loss $L_{cls}$ for the labeled set, along with a consistency loss $L_{cons}$ for the whole batch:
\begin{equation} \label{eq:loss}
\begin{aligned}
&\min_{\theta} \sum_{x_i \in (B \cap D_L)} L_{cls}(f(x_i,\theta, \eta), y_i)\\
&+  \lambda(e) \sum_{x_i \in B} L_{cons}\big( f(x_i,\theta', \eta') - f(x_i,\theta, \eta) \big), 
\end{aligned}
\end{equation}
where $L_{cls}$ is the standard cross-entropy loss, and $L_{cons}$ is the consistency loss to measure the distance between the softmax prediction of the teacher and the prediction of the student. We use the Mean Squared Error (MSE) loss: $L_{cons} = || f(x_i,\theta', \eta') - f(x_i,\theta, \eta) ||^2$. Here, ($\theta,\eta$) and ($\theta',\eta'$) represent the weights and perturbation parameters (e.g. augmentation and dropout) of the student and teacher models, respectively. $\lambda(e)$ is an epoch-dependent ramp-up weighting function, which controls the trade-off between the two loss terms. 

We maintain teacher's weights $\theta'$ as an EMA~\cite{tarvainen2017mean} of student's weights $\theta$ at training step $s$ as Eq.~\ref{eq:mt}. Note that the teacher model is directly generated using the EMA weights of the student model, and hence does not involve any training process.
\begin{equation} \label{eq:mt}
\theta'_{s} = \alpha \theta'_{s-1} + (1-\alpha) \theta_{s} \, ,
\end{equation}
where $\alpha$ is the smoothing hyperparameter called EMA decay, controlling the updating rate of the teacher. A small $\alpha$ enables large update of the teacher according to the student's weights at each step, resulting in a teacher of high similarity with the student. However, if the student learns too much unreliable targets, a large update can degrade the quality of the teacher. Mean teacher~\cite{tarvainen2017mean} address this problem by using a large $\alpha$ (e.g. 0.99) to have a teacher of low similarity with the student. Rather than relying on the hyperparameter, our more general solution is to improve the performance of the student by letting it learn from more reliable targets instead of uncertain noisy targets. Consequently, with a better student, the quality of the targets generated by the "shadow" model teacher can also be improved.

We introduce certainty-driven consistency on each training data in the loss term $L_{cons}$ in Eq.~\ref{eq:loss}. In particular, (1) Filtering CCL: the consistency loss is only computed on a subset certain targets, selected by our proposed uncertainty guided filtering algorithm (see Section~\ref{subsec:filter}) with a binary filtering mask $M$, using a mixture of two filtering strategies: hard filtering and probabilistic filtering. (2) Temperature CCL: we use a relative high temperature $V_i$ to reduce the loss magnitude of uncertain predictions vs. certain ones in the consistency loss term (see Section~\ref{subsec:temp}). An alternative is to use loss weight. However, this requires a deliberate design of suitable weights for all individual predictions.

\begin{figure}[t!]
    \centering
    \begin{subfigure}[t]{0.5\columnwidth}
        \centering		
        \includegraphics[width=.9\textwidth]{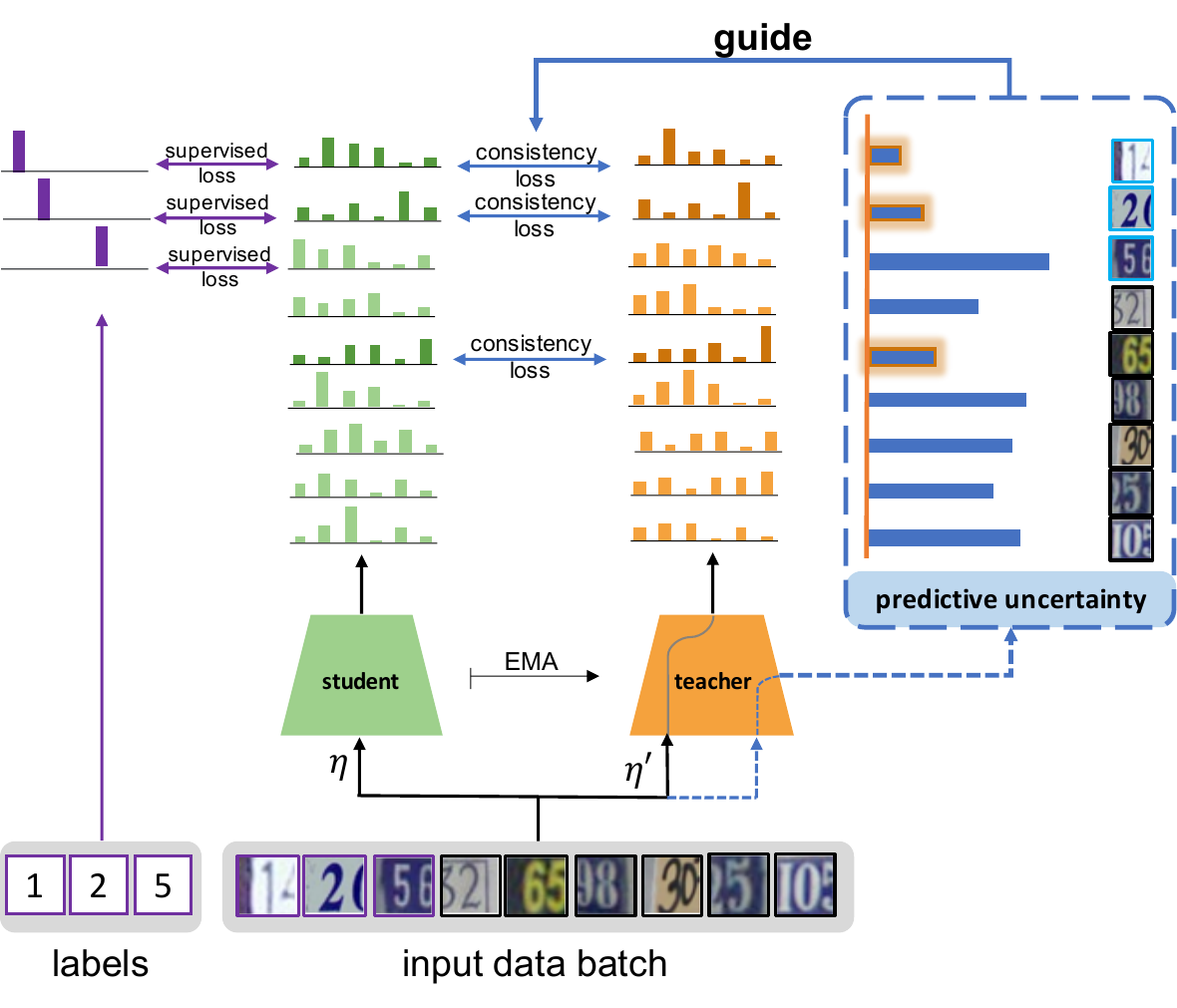} 
        \caption{Filtering CCL}
    \end{subfigure}%
    ~ 
    \begin{subfigure}[t]{0.5\columnwidth}
        \centering
        \includegraphics[width=.9\textwidth]{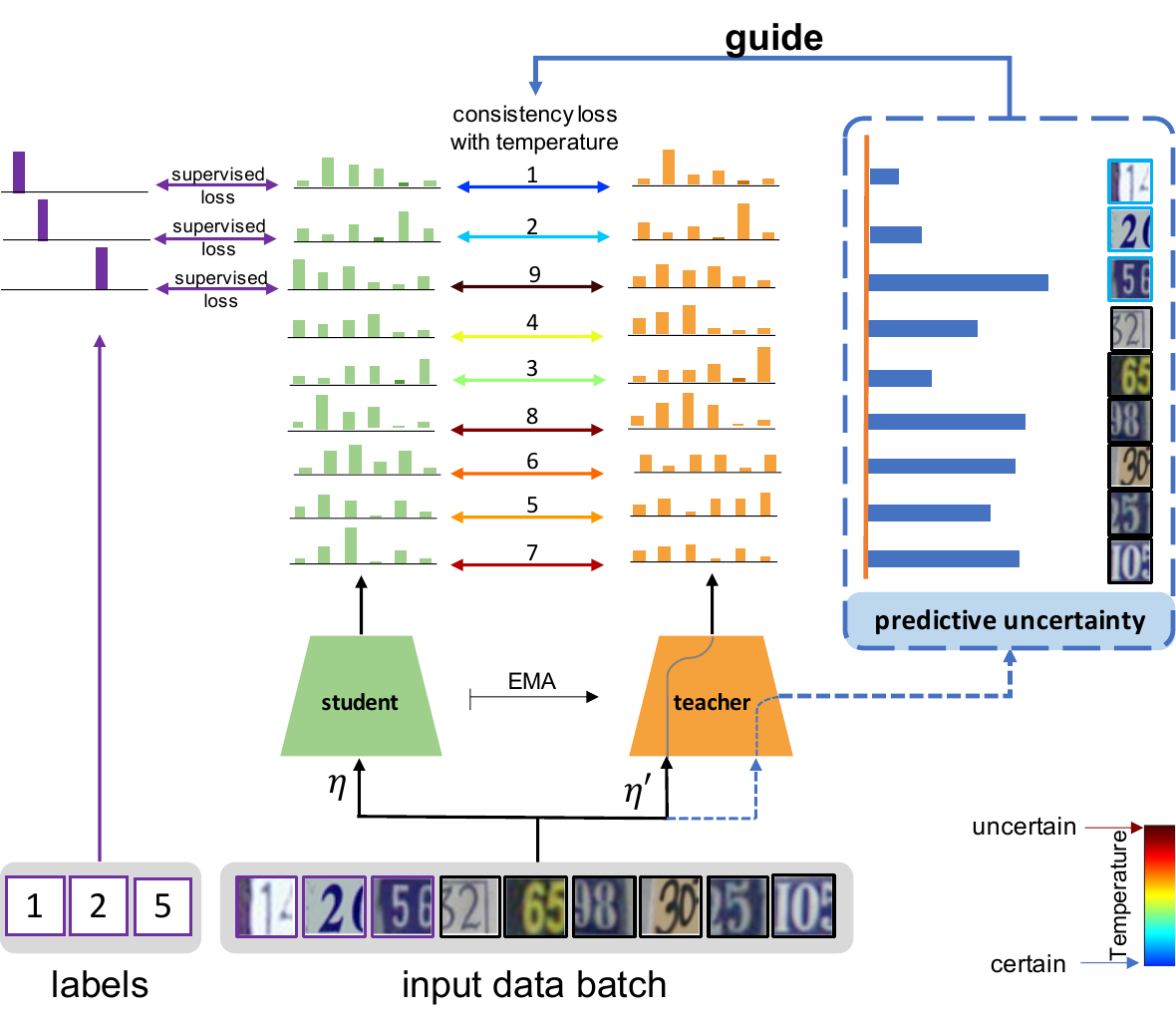}
        \caption{Temperature CCL}
    \end{subfigure} 
    \caption{(a) In Filtering CCL, the teacher filters out uncertain predictions and gradually selects a subset of certain predictions for the student to learn from. (b) In Temperature CCL, the teacher remains all targets but raises the temperature of the uncertain predictions, resulting a smaller loss on uncertain targets. Better viewed in color.}
\label{fig:network} 
\end{figure}

The framework of our proposed method is illustrated in Fig.~\ref{fig:network}. Given an input mini-batch, besides outputs predictions (i.e. softmax probability), the teacher evaluates the uncertainty of each of the predictions. We estimate the predictive uncertainty by measuring the variance or entropy of multiple predictions given by the up-to-date teacher model, under random input augmentations and dropout. In Filtering CCL, shown in Fig.~\ref{fig:network} (a), the teacher filters out uncertain predictions and gradually selects a subset of certain predictions (i.e. of low uncertainty), that are robust targets for the student to learn from. In Temperature CCL, shown in Fig.~\ref{fig:network} (b), the teacher remains all targets but raises the temperature of the uncertain predictions, resulting in a less penalty on uncertain targets if the student gives inconsistent predictions with the teacher. Both hard and soft CCL follow a ramp-up learning paradigm, that enables the student to gradually learn from relatively certain/easy to uncertain/hard cases. We believe that our more general solution can be more effective than most existing perturbation-based methods that ignore the quality of the loss of the individual training sample.

\begin{figure}[t!] 
         \centering 
         \includegraphics[width=1\columnwidth]{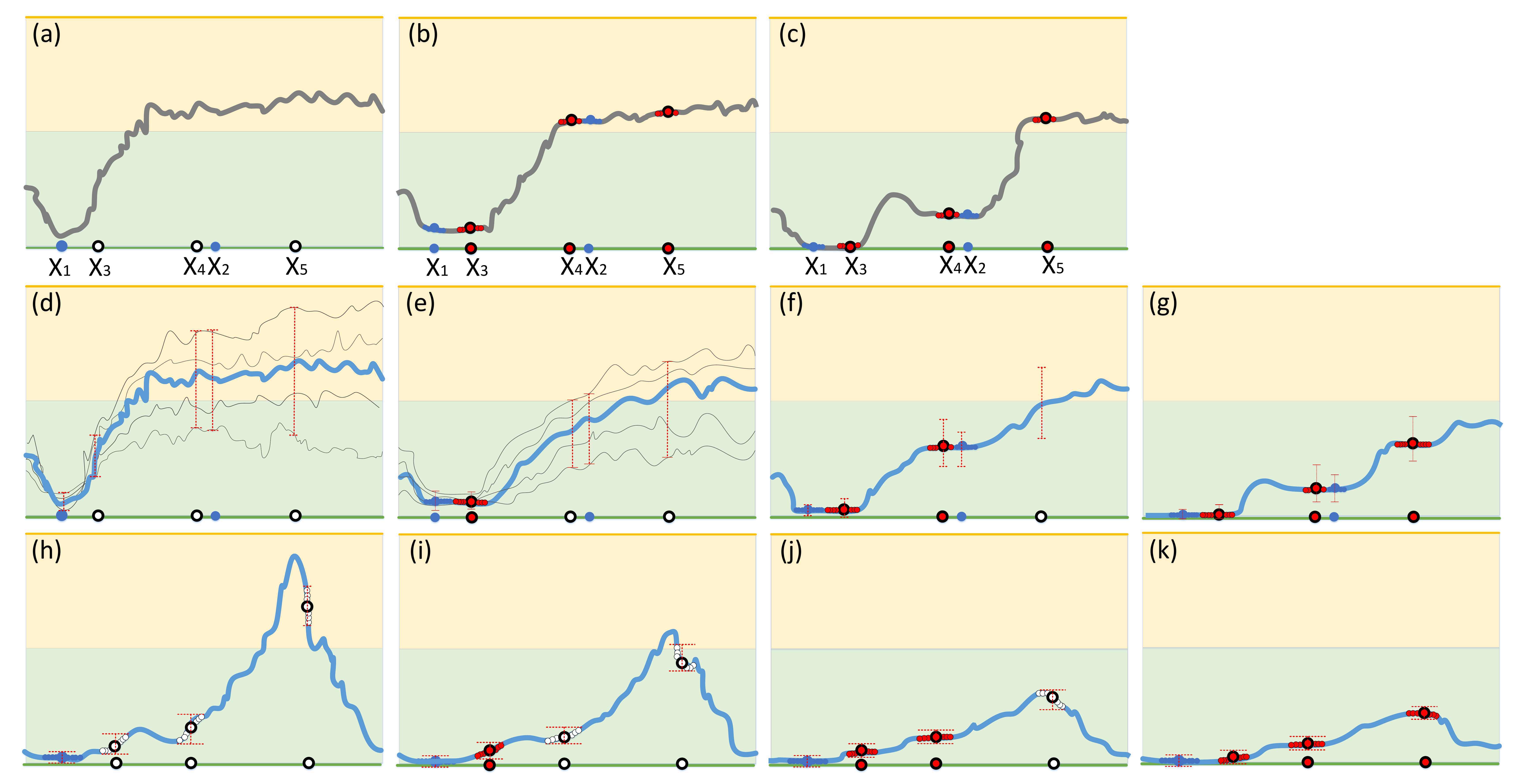}
         \caption{A sketch of a binary classification task with two labeled data $x_1,x_2$ and three unlabeled data $x_3,x_4,x_5$. Blindly penalizing the inconsistency around all data points can hinder learning (see (a-c)). Our approach dynamically selects (Filtering CCL) or pays more attention (Temperature CCL) to the certain predictions by estimating the model uncertainty under random dropout (d-g), and the local smoothness under input augmentations (h-k). A data on x-axis filled with red means it is used to enforce consistency. We omit the random sampled prediction curves in (f, g). Better viewed in color.}
        \label{fig:idea} 
\end{figure}

\subsection{Certainty-Driven Consistency Loss}
\label{sec:method}
Mean teacher~\cite{tarvainen2017mean} shows that the consistency of perturbated predictions around both labeled and unlabeled data provides smoothness regularization. However, in our view, it is an ideal case that all training data can be correctly classified. In reality, both labeled and unlabeled data can be misclassified. Penalizing inconsistency around misclassified data not only slow down the convergence of labeled data, but also can be harmful for the nearby unlabeled data. Without ground-truth supervision, the issue of the confirmation bias can make the model stuck in these incorrect inconsistency constraints, and can hardly be revised in the successive training process. Blindly enforcing smoothness around all samples can induce the confirmation bias. Although in the next training epoch, the supervised loss can correct the prediction of a labeled data, some unlabeled data can still remain in the previously enforced wrong prediction, given that there is no ground-truth label available for it.

Our main idea is not to simply just train the teacher for doing predictions, but also to estimate the underlying uncertainty associated with them, so that it can then gradually select the most certain predictions as targets for the student to learn from. As illustrated in Fig.~\ref{fig:idea}, we perform two kinds of perturbations to estimate uncertainty: dropout (Fig.~\ref{fig:idea} (d-g)), and random input augmentation (Fig.~\ref{fig:idea} (h-k)). The former estimates model uncertainty~\cite{gal2016dropout}, and the latter estimates local smoothness. Taking Filtering CCL as a example, in the initial training epochs, the supervised classification loss can be high, and the predictions can vary considerably under different perturbations (Fig.~\ref{fig:idea} (d, h)). In this case, the teacher can only select a small number of high certainty targets with relatively low variances to enforce consistency (red dots in Fig.~\ref{fig:idea} (e, i)). As training continues, the overall classification loss decreases and so does consistency loss. The uncertainty level of the teacher predictions can be reduced gradually, allowing the student to learn from more reliable targets of unlabeled data (Fig.~\ref{fig:idea} (f, g, j, k)). As for Temperature CCL, the red dots means they have higher impact (lower temperature) in the consistency loss compared to uncertain ones (white).

\subsection{Uncertainty Estimation}
\label{subsec:metric}
Given a batch of input training images containing both labeled and unlabeled data, each training step includes two stages: prediction and uncertainty estimation. In the prediction stage, the student and teacher output two sets of predictions respectively. Existing methods directly use the teacher's predictions as targets for the student to learn from. Instead, we introduce uncertainty estimation to dynamically select a subset of certain predictions that are associated with reliable targets to enforce consistency constraints. In the uncertainty estimation stage, we perform $T$ stochastic forward passes on teacher model under random dropout $\hat{\theta'^t}$ and input augmentation $\hat{\eta'^t}$.

For each input data $x$ at $t$th forward pass, we obtain a softmax probability vector $[p(y=1|x, \hat{\theta'^t},\hat{\eta'^t}), ..., p(y=C|x, \hat{\theta'^t},\hat{\eta'^t})]$. Collecting a set of $T$ predictions for each $x$, we are able to estimate the teacher model's predictive uncertainty $U$. This can be considered as the Monte Carlo sampling from the posterior distribution of models~\cite{gal2016dropout}. Different from~\cite{gal2016dropout} which only measures model uncertainty by random dropout, our uncertainty also takes the local smoothness into consideration. Our key reasoning is that a certain prediction should meet two requirements: to predict consistently for the same input using randomly sampled sub-networks, and to predict consistently for the similar input pair with randomized augmentations.

We investigate four metrics to approximate uncertainty: predictive variance (PV), entropy variance (EV), predictive entropy (PE)~\cite{freeman1965elementary, Gal2016Uncertainty}, and mutual information (MI)~\cite{shannon1948mathematical, Gal2016Uncertainty}. The criteria of choosing a metric is: (i) It can measure the variance over $T$ times random samplings. (ii) It can reflect the probability distribution of different classes, rather than representing a hard prediction, i.e. the top one predicted class (argmax). Hence, we do not use predictive ratio (PR)~\cite{Gal2016Uncertainty} which is the frequency ($t_{mode}$) of the mode predicted class ($mode$) over $T$ times: $PR = 1 - t_{mode}/T$. (iii) It gives a continuous scalar value in order to compare the uncertainty of different data points with high precision. 

Specifically, (1) PV has been used in regression task~\cite{gal2016dropout}. We apply PV in classification task to measure the variance of multiple soft predictions obtained from stochastic forward passes for all classes. The larger variance with respect to the $T$ times mean, the higher uncertainty. (2) EV is the variance of $T$ times' entropies of the predictions. If the entropy $H$ varies a lot over $T$ times, we consider the model is uncertainty. (3) PE~\cite{freeman1965elementary,Gal2016Uncertainty} captures the entropy of the averaged probability distribution over $T$ times. It attains its maximum value when all classes are predicted to have equal uniform probability, and its minimum value of $0$ when one class has probability $=1$ and all others probability $=0$, i.e. a certain prediction. (4) MI~\cite{shannon1948mathematical,Gal2016Uncertainty} equals to PE minus the average entropy over $T$ times stochastic passes, which combines the entropy of the expected prediction with each prediction's entropy. Overall, by measuring the uncertainty of $T$ times soft predictions, we effectively estimate how close the $T$ times distributions is to each other. This can be seen as the uncertainty of "dark knowledge"~\cite{hinton2015distilling}, which is a much stronger estimation compared to measuring whether only the final classification remains the same. 
The respective definitions of the metrics are as follows:
\begin{equation} \label{eq:metric}
\begin{aligned}
&\mu_c \, \,= \frac{1}{T} \sum_t p(y=c|x, \hat{\theta'^t}, \hat{\eta'^t})\\
&H_t \, \,= \sum_{c} p(y=c|x, \hat{\theta'^t}, \hat{\eta'^t}) \log p(y=c|x,\hat{\theta'^t}, \hat{\eta'^t})\\
&PV = \sum_c \text{Var}[p(y=c|x, \hat{\theta'^1}, \hat{\eta'^1}), ..., p(y=c|x, \hat{\theta'^T}, \hat{\eta'^T})] \\
\, &\, \, \, \quad = \sum_c \Big( \frac{1}{T} \sum_t (p(y=c|x, \hat{\theta'^t}, \hat{\eta'^t}) - \mu_c)^2 \Big)\\
&EV = \text{Var}[H_1, ..., H_T]\\
&PE = - \sum_c \mu_c \log \mu_c\\
&MI = PE - \frac{1}{T} \sum_t H_t
\end{aligned}
\end{equation}

In practice, in the prediction stage, we turn off dropout inside the teacher and use the whole network to provide a robust prediction target, which is the expected output over the previous training steps. In the uncertainty estimation stage, we turn on dropout in the teacher network to obtain multiple randomly connected sub-networks. For the student, since it is trained with loss back-propagation, we keep the dropout on throughout training to prevent over-fitting, as done in the standard dropout procedure~\cite{srivastava2014dropout}.

\subsection{Certainty-Driven Consistency with Filtering}
\label{subsec:filter}

\begin{figure} 
	\centering 
	\includegraphics[width=0.7\columnwidth]{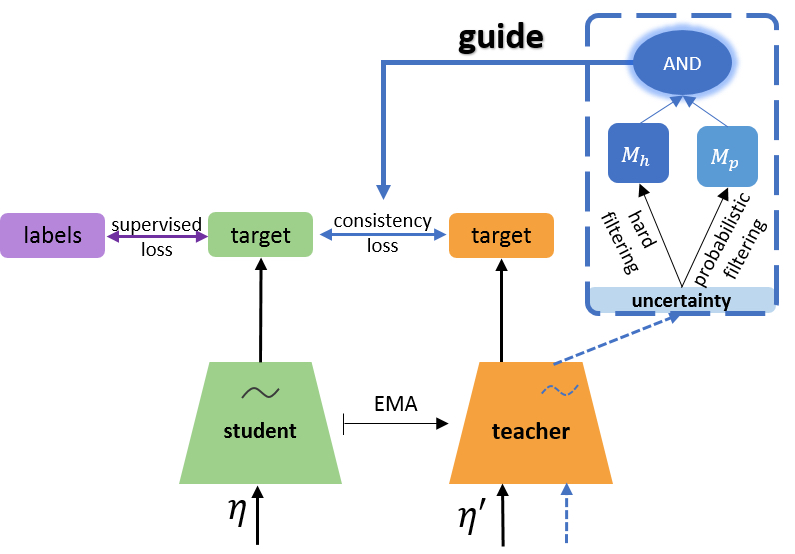}
	\caption{Filtering strategies in our Filtering CCL.}
	\label{fig:filter} 
\end{figure}

Using the mentioned uncertainty measurement, the teacher outputs an uncertainty value $U_i$ for each data points $x_i$ in the input batch, along with their soft targets. Based on $U_i$, we explore two filtering strategies -- \textit{hard filtering} and \textit{probabilistic filtering} to filter out relatively uncertain predictions when computing the consistency loss  (Fig.~\ref{fig:filter}). \textit{Hard filtering} ensures that the student always learns from the targets that are of relatively high quality, i.e. with low predictive uncertainty. \textit{Probabilistic filtering} filters a data sample by a probability value related to its uncertainty ranking. A sample with high uncertainty has high probability to be filtered out, but still has a chance to remain. This strategy introduces complementary randomness into the filtering process, which can improve the generalization performance.

In each training step, the data samples in the input batch $B$ are sorted according to their uncertainty values $[U_1,...,U_{|B|}]$ in ascending order to obtain an ordered rank list $R=[R_1,...,R_{|B|}]$. The first on $x_i$ is the most certain in the batch, i.e. $R_i = 1$, and the last one $x_j$ is the most uncertain, i.e. $R_j = |B|$. (1) Hard filtering: We choose a ramp-up number of top-$k$ certain data points and filter out the left uncertain data, where $k=\beta e$, $e$ is the epoch. The hard filtering mask is denoted as $M_h$ (length of $B$), and for input $x_i$, ${M_h}_i=1$ if $R_i \le k$, otherwise ${M_h}_i=0$. (2) Probabilistic filtering: The probabilistic filtering mask is denoted as $M_p$. Each element of $M_p$ is a Bernoulli distributed random variable ${M_p}_i=\{0,1\}$ expressed as:
\begin{eqnarray} \label{eq:filter}
{M_p}_i &\sim \text{Bernoulli}({m_p}_i), 
\end{eqnarray}
where:
\begin{eqnarray}
\nonumber
{m_p}_i&=&\frac{P_{max}}{|B|-1} (R_i - 1),\\ \nonumber
P_{max}&=&\left\{
\begin{array}{l}
1- \rho  \frac{e}{E} \, , \, if \, e \le E\\
\quad 0 \,\,\,\, \quad, \, if \, e > E,
\end{array}
\right.
\end{eqnarray}
where $\rho \in (0,1)$ is a coefficient hyperparameter influencing the value of $P_{max}$, and $E$ is a thresholding epoch deciding from which epoch we want to exploit all training data in the consistency loss.

We find that it is more stable to use this mapped value based on rank $R_i$ as the filtering probability $m_{p_i}$ compared to the original uncertainty value $U_i$. In particular, for the most certain data $R_i=1$ in the batch, the probability of being filtered out is always $0$ (${m_p}_i=0$). For the most uncertain data $R_j=|B|$,  the probability of filtering it out equals to $P_{max}$, which decreases as the training epoch increases. For the data in between these two extremes, the probability of being filtered out follows a uniform distribution ranging from $0$ to $P_{max}$. The reason we use this mapped value is that an absolute uncertainty value $U_i$ can be numerically small but relatively larger than that of other data in the same batch or other batches. In this case, we still want to filter it out with a relatively high probability. Additionally, non-uniformly distributed filtering can cause a large variation in terms of the number of the selected data among the batches, which can make the training process unstable. 

Our method can gradually select the most certain predictions as the targets for the student to learn from. In the initial training epochs, the supervised classification loss can be high, and the predictions can vary considerably under different perturbations. In this case, the teacher can only select a small number of high certainty targets with relatively low variances to enforce consistency. As training continues, the overall classification loss decreases and so does the consistency loss. The uncertainty level of the teacher predictions can be reduced gradually, allowing the student to learn from more reliable targets of unlabeled data.

Note that, we do not explicitly differentiate between labeled and unlabeled data when computing our consistency loss. As mentioned before, the consistency constraints of both the misclassified labeled and unlabeled data can be harmful to learning convergence and generalization performance. In practice, since the learning curve of the labeled data will converge quickly with supervised classification loss, the average uncertainty level of the labeled data is usually lower than the unlabeled data. Also, as training continues, both the student and the teacher can learn more and more reliable knowledge, and the system generates more and more stable predictions instead of just noisy random guesses, which can then reduce the overall uncertainty level of the model. Hence, our filtering strategy automatically retains more and more labeled data, and filter out less and less uncertain data.

\subsection{Certainty-Driven Consistency with Temperature}
\label{subsec:temp}
Besides filtering out uncertain targets in a hard way, we also investigate a soft way by letting the student pay less attention on learning from such uncertain targets. We adopt the temperature softmax function~\cite{sutton1998introduction} into our consistency regularization to adjust the magnitudes of the gradients of the training samples with various uncertainty values. 

The temperature function is a variant of the softmax activation function by dividing the logit $z_i$ for class $i$ with a positive temperature $V$:
\begin{equation} \label{eq:temp}
\begin{aligned}
q_i = \frac{\exp(z_i/V)}{\sum_j \exp(z_j/V)}, 
\end{aligned}
\end{equation}
where $q_i$ is the output of the temperature softmax for class $i$. A higher temperature $V$ produces a softer probability distribution, i.e. all classes are equally distributed. A lower temperature $V$ causes all classes to be sparsely distributed.

The scaling effect in the temperature softmax is discussed by Hinton et al.~\cite{hinton2015distilling} in the context of model distillation. It has been theoretically proved that in the high temperature limit, minimizing the consistency loss between two predictions with temperature $V$ has the effect of reducing the magnitudes of the gradients by a scale of $1/V^2$. However, ~\cite{hinton2015distilling} needs to manually set the value of $V$, which is the same for all training samples. On the contrary, our method can automatically define the temperature of each training sample according to its uncertainty. We use different temperatures to control the magnitude of the consistency loss to let the student learn more from certain targets and less from uncertain ones. We apply temperature softmax to both student and teacher models.

Specifically, we use higher temperatures to reduce the magnitudes of the gradients coming from uncertain targets, to let the student learn more from certain targets and less from uncertain ones. In each training step, we automatically enforce relatively higher temperatures on uncertain targets, and lower temperatures on certain targets, according to the previously obtained certainty ranking list $R$ in each input batch $B$:
\begin{equation} \label{eq:temp_uncert}
\begin{aligned}
V_i = (\frac{R_i}{|B|})^2 V_{max} + 1,\quad V_{max} = \text{max}(V_b - \frac{e}{E}), 1),
\end{aligned}
\end{equation}
where $V_i$ is the temperature for the input $x_i$, $|B|$ is batch size, $R_i$ is the certainty rank of $x_i$, $V_{max}$ is the maximum temperature value for the most uncertain data in each batch, $V_b$ is the base temperature, and $e$ is the current training epoch. We use quadratic function to make the temperature values of certain and uncertain data in the batch differentiable. $V_{max}$ decreases as training continues, since the system is more and more certain. $E$ is a threshold epoch controlling the speed of the ramp-down function.

Besides the scaling effect, our temperature CCL also encourages diverse gradients coming from various classes rather than the predicted (i.e. argmax) class for the uncertain targets. Our idea is, for certain targets, we can trust more in the argmax class compared to uncertain ones, since the argmax prediction is probably correct. However, for the uncertain targets, it can be dangerous to focus only on the error-prone argmax class and ignore other classes. Hence, we increase the temperature of uncertain data to soften the probability distribution, so that the influence from different classes becomes more balanced. This balanced distribution can reduce the total amount of the unreliable error-prone gradients in the loss back-propagation procedure.

\subsection{Combining Filtering and Temperature CCL}
We further analyze the effectiveness of combining our proposed Filtering CCL and Temperature CCL, denoted as FT-CCL. The motivation is that, with regard to learning from certain targets, the two approaches have complementary effects.  In particular, in each training step, according to the previously obtained certainty ranking list $R$ for each input batch $B$, we obtain the two masks $M_h$ and $M_p$ for hard filtering and probabilistic filtering respectively, and the temperature value list. 

We first filter out uncertain predictions based on $M_h$ and $M_p$, and then, compute the outputs of the temperature softmax for the remaining certain predictions using their corresponding temperature values. In the initial training epochs, the Filtering CCL plays a dominant role, since it only select a few top-$k$ certain predictions for the consistency. In the following epochs, Temperature CCL becomes more and more important, since the temperature values of the remaining predictions becomes more and more differentiable. Yet, as training continues, such differences reduces since the overall certainty level increases.

\subsection{Decoupled Multi-Teacher}

The typical perturbation-based methods~\cite{lee2013pseudo,rasmus2015semi,laine2016temporal,tarvainen2017mean} use one single teacher to provide a prediction target for a student to learn from. In MT~\cite{tarvainen2017mean}, the teacher produced by the student also teaches the same student in terms of the consistency. This strongly coupled student-teacher can limit the capacity of the model, due to the high similarity between them. 
Aiming to improve the generated targets, we extend the single student-teacher framework to a mutually learning framework containing pairs of students and teachers. 

\begin{figure} 
	\centering 
	\includegraphics[width=0.7\columnwidth]{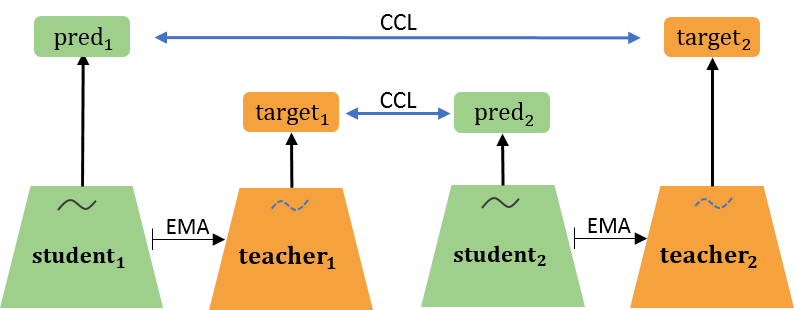}
	\caption{Our proposed circle-shape framework with two decoupled students and teachers.}
	\label{fig:4view} 
\end{figure}

It has been shown that mutual learning between two models with different initialization can provide additional knowledge to each other, and hence improves performance for model distillation in fully supervised scenario~\cite{zhang2018deep}. The key condition of mutual learning~\cite{zhang2018deep} to be successful is that, the model from different initialization can provide additional information about each data. We introduce a mutual learning framework to decouple the students and teachers, by forming them in a closed circle (Fig.~\ref{fig:4view}). 
Assume we have a sequence of $n$ pairs of student-teacher. The $student_i$ produces an EMA model as $teacher_{i}$, which teaches the next $student_{i+1}$, and the last $teacher_{n}$ teaches the first $student_1$, forming a circle of a learning process.
In this way, the teacher does not directly teach the student that generated it anymore, but teaches the next student in a circle manner. Since the feedback connection between a student model and its EMA-self (i.e. teacher) has to go through an intermediary (i.e. a circle), one can consider the EMA duration has been extended, encouraging the individual models to be diverse. To further avoid the models collapsing into each other, we apply different input augmentations randomly for all students/teachers.

Note that, in order to prove the effectiveness of our circle-shape framework, we do not ensemble the multiple teacher models for the final evaluation. Instead, we only report the average performances of these models.

\section{Experiments}

We show the experimental results on three widely adopted image classification benchmark datasets: SVHN~\cite{netzer2011reading}, CIFAR-10 and CIFAR-100~\cite{krizhevsky2009learning}. The Street View House Numbers (SVHN) includes 32 $\times$ 32 RGB images of real-world house numbers (0~9), and the task is to classify the centermost digit. It consists of 73,257 training images and 26,032 test images. The CIFAR-10 consists of 32 $\times$ 32 natural images from 10 classes such as airplanes, cats, and dogs. It contains 50,000 training images and 10,000 test images. The CIFAR-100 dataset consists of 50,000 training images and 10,000 test images from 100 more fine-grained object classes.

Following the standard semi-supervised classification protocol~\cite{rasmus2015semi,salimans2016improved,laine2016temporal,tarvainen2017mean,luo2018smooth}, we randomly sample 1000, 4000 and 10000 labels for SVHN, CIFAR-10 and CIFAR-100, respectively, with the remaining 72257, 46000, 40000 training images as unlabeled training data. The results are averaged over 5 runs with different seeds for data splits.

\subsection{Implementation Details} 
We implemented our code using PyTorch~\cite{paszke2017automatic}. We adopt the 13-layer CNN architecture as~\cite{laine2016temporal, tarvainen2017mean}. We initialize the network weights using a self-supervised pretraining (i.e. RotNet~\cite{gidaris2018unsupervised}), which is to predict the image rotations ($\{0^{\circ}, 90^{\circ}, 180^{\circ}, 270^{\circ}\}$). In the uncertainty estimation analysis, we initialize the network from random weights to better show the effectiveness of our uncertainty estimation. Following the previous works~\cite{laine2016temporal, tarvainen2017mean}, we use random translation ($[-2, 2]$ pixels) on SVHN and horizontal flips and random translation on CIFAR-10 and CIFAR-100.  We train the network with minibatches of size 512, including 128 labeled and 384 unlabeled data. We use SGD optimizer for training with base learning rate 0.1, a weight decay of $2e-4$, and a momentum of 0.9. The teacher model weights are updated after each training step using an EMA with $\alpha = 0.99$. Following~\cite{tarvainen2017mean,laine2016temporal}, we ramped up the consistency loss coefficient $\lambda$ during the first 80 epochs using a sigmoid-shaped function $\text{exp}[-5(1-x)^2]$, where $x$ advances linearly from zero to one during the ramp-up period. We run 10 times dropout during training to measure uncertainty, which we find is sufficient for our purpose of estimating uncertainty. Except PR, the four uncertainty metrics (PV, EV, PE, MI) show similar performance. Unless specified, we use PV as the default uncertainty metric. In Filtering CCL, hard filtering strategy selects top-$k$ certain predictions using a linear ramp-up function $k=8e$, where $e$ is epoch, and probabilistic filtering sets the maximum masking probability as $P_{max}(e)=1-0.4e/210$. In Temperature CCL, the maximum temperature $V_{max}(e)=4-e/80$ bounded by 1. These hyper-parameters are set through grid-search on the validation set. For fair comparison, in our decoupled multi-teacher method, we do not ensemble the multiple teacher models. Instead, we only report the average performances of these models.

\begin{table*}[ht!]
\scriptsize
	\caption{Test error rate on benchmark datasets, averaged over 5 runs. \textbf{Metric}: Error rate (\%) $\pm$ standard deviation, \textbf{lower is better}. "--" indicates no reported result, "*" is based on our best implementation, "1" result is reported in~\cite{laine2016temporal}.}
	\label{tab:svhn_cifar10_cifar100}
	\centering
	\begin{tabular}{l | ccc|c |c}
		\toprule
		Model &  &  CIFAR-10 &  & CIFAR-100 & SVHN  \\
		&1000 labels & 2000 labels & 4000 labels
		& 10000 labels
		& 1000 labels 
		\\
		\midrule
		Supervised-only~\cite{tarvainen2017mean}
		&46.43 $\pm$ 1.21&33.94 $\pm$ 0.73&20.66 $\pm$ 0.57
		&\,\,44.56 $\pm$ 0.30\footnotemark
		&12.32 $\pm$ 0.95 
		\\
		$\Pi$ model~\cite{laine2016temporal} 
		&--&--&12.36 $\pm$ 0.31
		&39.19 $\pm$ 0.36
		&4.82 $\pm$ 0.17
		\\
		TempEns~\cite{laine2016temporal} 
		&--&--&12.16 $\pm$ 0.24
		&38.65 $\pm$ 0.51
		&4.42 $\pm$ 0.16
		\\
		VAT+Ent~\cite{miyato2018virtual} 
		&--&--&10.55 $\pm$ 0.05
		&--
		&3.86 $\pm$ 0.11
		\\
		MT~\cite{tarvainen2017mean} 
		&21.55 $\pm$ 1.48&15.73 $\pm$ 0.31&12.31 $\pm$ 0.28
		&\,\,37.91 $\pm$ 0.37*
		&3.95 $\pm$ 0.19
		\\
		$\Pi$+SNTG~\cite{luo2018smooth}
		&21.23 $\pm$ 1.27&14.65 $\pm$ 0.31&11.00 $\pm$ 0.13
		&37.97 $\pm$ 0.29
		& 3.82 $\pm$ 0.25
		\\
		MT+SNTG~\cite{luo2018smooth}
		&--&--&--
		&--
		&3.86 $\pm$ 0.27
		\\
		TempEns+SNTG~\cite{luo2018smooth}
		&18.41 $\pm$ 0.52&13.64 $\pm$ 0.32&10.93 $\pm$ 0.14
		&--
		&3.98 $\pm$ 0.21
		\\
		MA-DNN~\cite{Chen2018ECCV}
		&--&--&11.91 $\pm$ 0.22
		& 34.51 $\pm$ 0.61
		& 4.21 $\pm$ 0.12
		\\
		LP~\cite{2019dlp}
		&16.93 $\pm$ 0.7 &13.22 $\pm$ 0.29&10.61 $\pm$ 0.28
		& 35.92 $\pm$ 0.47
		& --
		\\
		Co-Train~\cite{qiao2018deep}
		& -- & -- & 9.03 $\pm$ 0.18
		& 34.63 $\pm$ 0.14
		& 3.61 $\pm$ 0.15
		\\
		MT+FSWA~\cite{athiwaratkun2019there}
		& 15.58 $\pm$ 0.12 & 11.02 $\pm$ 0.23 & 9.05 $\pm$ 0.21
		&  33.62 $\pm$ 0.54
		& --
		\\
		WCP~\cite{zhang2020wcp} 
		&  17.62 $\pm$ 1.52 &  11.93 $\pm$ 0.39 & 9.72 $\pm$ 0.31
		& --
		& 3.58 $\pm$ 0.18
		\\
		DS~\cite{ke2019dual}
		& 14.17 $\pm$ 0.38 & 10.72 $\pm$ 0.19 & \textbf{8.89 $\pm$ 0.09}
		& \textbf{32.77 $\pm$ 0.24}
		& -- 
		\\
		\midrule \midrule
		Filtering CCL
		& 14.35 $\pm$ 0.51& 11.76 $\pm$ 0.34 & 9.77 $\pm$ 0.16
		& 34.07 $\pm$ 0.47
		& 3.70 $\pm$ 0.23
		\\
		Temperature CCL
		&14.48 $\pm$ 0.59 & 11.84 $\pm$ 0.33 & 9.90 $\pm$ 0.21
		&34.19 $\pm$ 0.42
		&3.73 $\pm$ 0.16
		\\
		FT-CCL
		&14.14 $\pm$ 0.46 &11.03 $\pm$ 0.24 & 9.65 $\pm$ 0.17
		&33.92 $\pm$ 0.36
		&3.67 $\pm$ 0.13
		\\ 
		FT-CCL with 2 T
		&13.68 $\pm$ 0.38 &10.59 $\pm$ 0.23 & 9.17 $\pm$ 0.19
		&33.51 $\pm$ 0.31
		&3.53 $\pm$ 0.18
		\\
		FT-CCL with 3 T
		&\textbf{13.45 $\pm$ 0.35} & \textbf{10.32 $\pm$ 0.21} & \textbf{8.89 $\pm$ 0.15}
		& 33.34 $\pm$ 0.32
		& \textbf{3.50 $\pm$ 0.17}
		\\
		\bottomrule
	\end{tabular}
\end{table*}

\subsection{Comparisons} 
We compare our proposed method with the existing methods in literature. Tab.~\ref{tab:svhn_cifar10_cifar100} shows the results of our proposed CCL approaches and the existing methods on the three benchmarks for the cases where different number of labels are given. The accuracy of existing methods are all taken from existing literature. In general, both of our Filtering CCL and Temperature CCL perform well, and our combined FT-CCL with multiple teachers further boosts the performance across all experimental settings on three datasets. 
Among existing approaches, $\Pi$ model~\cite{laine2016temporal}, TempEns~\cite{laine2016temporal}, and MT~\cite{tarvainen2017mean} are three base methods which generate the teacher model by copying the student model, ensemble the student predictions, and ensemble the student model weights, respectively. The recently proposed DS~\cite{ke2019dual} jointly trains two student models together with the EMA teacher, and enforces the consistency loss on the stable samples.
DS~\cite{ke2019dual} achieves 14.17\% and 10.72\% error rate on CIFAR-10 with 1000 and 2000 labels, respectively. On CIFAR-10, our method achieves better performance compared to DS method, especially when labels are fewer. In particular, the test error rate of our decoupled FT-CCL with 3 teachers reaches 13.45\% and 10.32\% with 0.72 and 0.4 absolute gains compared to DS on CIFAR-10 with 1000 and 2000 labels, respectively. With 4000 labels, our decoupled FT-CCL with 3 teachers (8.89\%) performs comparably to DS (8.89\%).
Co-Training~\cite{qiao2018deep} exploits adversarial examples to improve teacher models.  MT+FSWA~\cite{athiwaratkun2019there} improves over MT using fast stochastic weight averaging to generate a stronger ensemble teacher. 
On CIFAR-100, which is a more challenging dataset with 100 number of classes, our proposed FT-CLL with 3 teachers achieves better performance (33.34\%) than the previous method Co-Training (34.63\%) and MT+FSWA (33.62\%).
The recently proposed WCP~\cite{zhang2020wcp} stabilizes the network predictions in presence of the worse-case perturbations imposed on the network weights and dropout structures.
On SVHN, our decoupled FT-CCL with 3 teachers achieves an error rate of 3.50\%, which outperforms the previous WCP method (3.58\%).

\begin{figure} 
	\centering 
	\includegraphics[width=0.7\columnwidth]{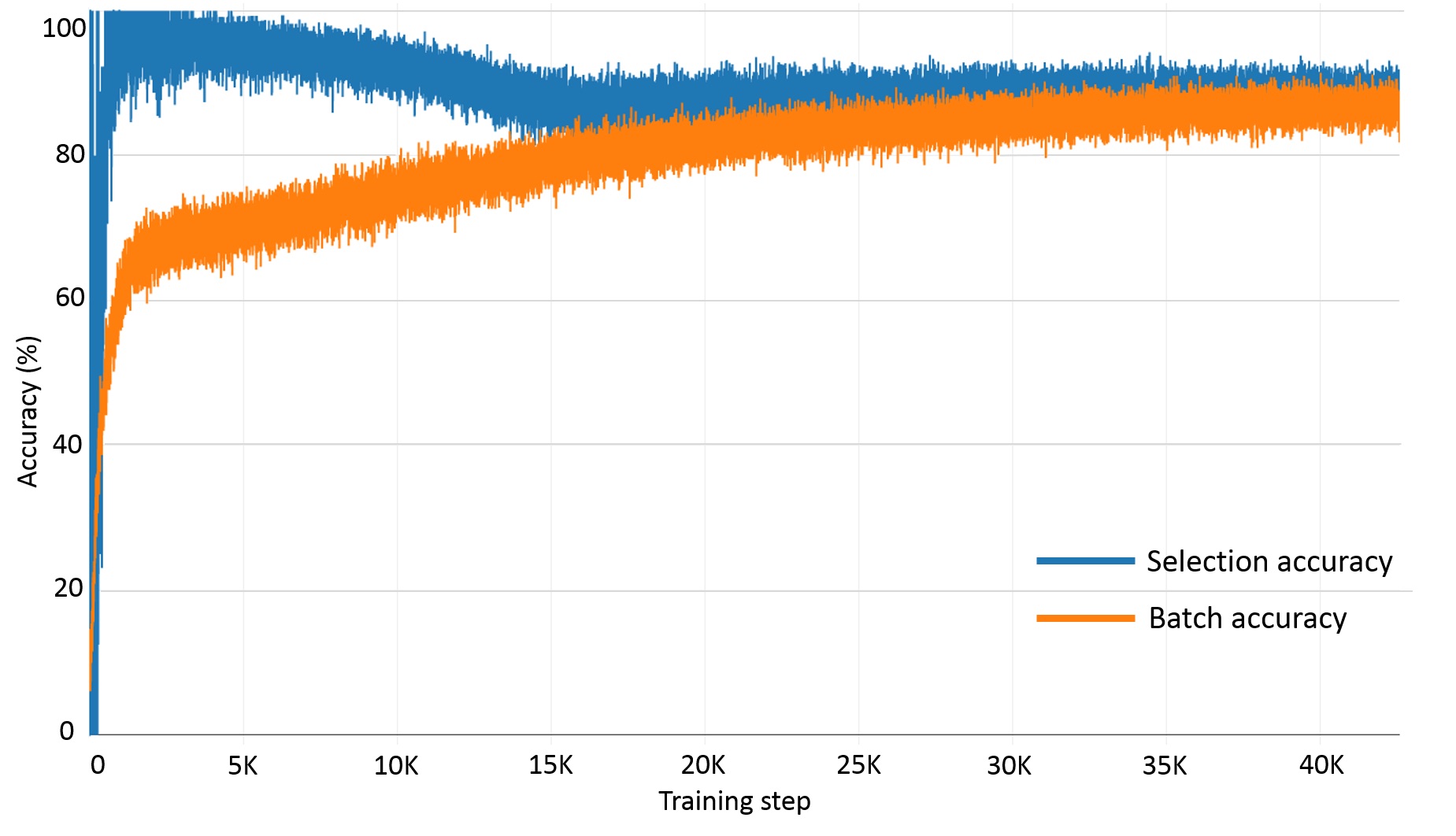}
	\caption{The prediction accuracy of our selected training samples with high certainty during training (blue) on CIFAR-10 with 1000 labels. Better viewed in color.}
	\label{fig:topk} 
\end{figure}

We further compare the computation and memory usage between our decoupled multi-teacher networks with the existing single teacher methods and co-training methods. Compared to single student-teacher methods, e.g. MT~\cite{tarvainen2017mean}, our decoupled 2 teacher model contains one more trainable student model, and one more EMA teacher model. Hence, the computation and memory cost can be doubled. Compared to co-training~\cite{qiao2018deep}, with the same number of trainable models (2 students), our method obtains better teachers ``for free" thanks to the EMA weights, which does not involve any optimization process.

\subsection{Effectiveness of Our Uncertainty Estimation} 
A potential concern of our Filtering CCL is the quality of the selected training samples during training, since the uncertainty estimation is likely to be unreliable at the beginning of the training process. Fig.~\ref{fig:topk} shows the prediction accuracy of our selected training samples with high certainty during training (blue), and the accuracy of all training samples in the input batch (orange). Note that, the ground-truth labels for unlabeled data are only used to evaluate the quality of our filtering algorithm, and have not been used for training. In the initial training step, although the selection accuracy fluctuates, since hard filtering only allows few (e.g. 8, 16) samples to be picked, the selection accuracy increases fast. As training continues, since we select more and more training data in the consistency loss, the two accuracy curve will merge together. The key that ensures the positive effect of our certainty driven consistency on convergence is our filtering strategy. Even though the estimated uncertainty can be wrong in the beginning, which can influence the probabilistic filtering performance, the hard filtering strategy only allows the teacher to slowly select the most certain data to enforce consistency. This dynamic certainty-driven and gradual ramp-up top-$k$ selection strategy allows the system to warm up.

\begin{figure}
    \centering
    \begin{subfigure}[t]{0.5\columnwidth}
        \centering		
        \includegraphics[width=1\columnwidth]{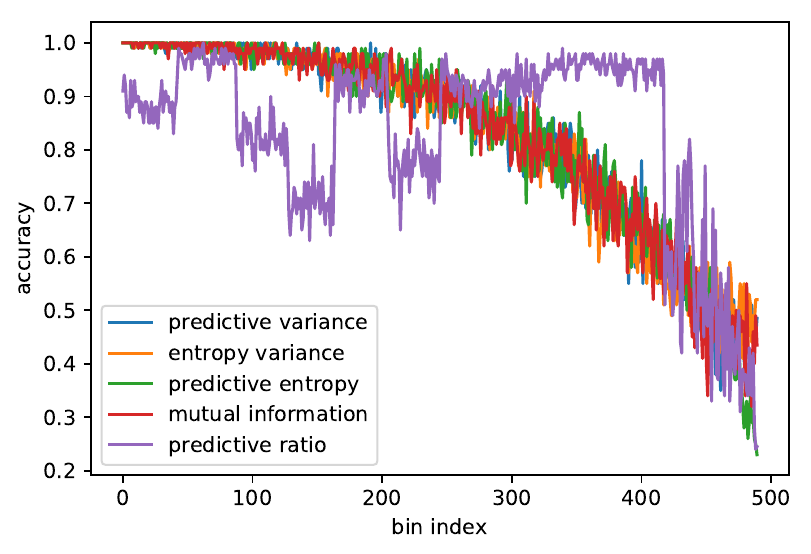}
        \caption{Accuracy of metrics on $D_U$}
    \end{subfigure}%
    ~ 
    \begin{subfigure}[t]{0.5\columnwidth}
        \centering
        \includegraphics[width=1\columnwidth]{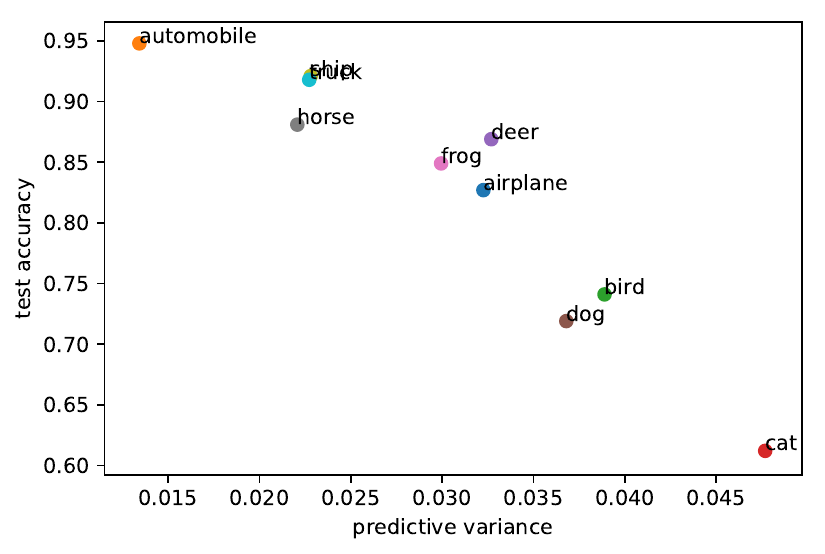}
        \caption{Accuracy vs. PV on $D_T$}
    \end{subfigure} 
    \caption{The inverse correlation between accuracy and uncertainty on CIFAR-10 trained with 1000 labels and 49000 unlabeled data ($D_U$). $D_T$ denotes test set. (a) Except predictive ratio, other four metrics give similar trends: higher uncertainty, lower accuracy. (b) shows the inverse relationship between class accuracy and predictive variance. Better viewed in color.}
\label{fig:var} 
\end{figure}

To further verify the effectiveness of our uncertainty estimations and understand the behaviour of different uncertainty metrics, we plot the relationship between accuracy (y-axis) and uncertainty estimates (x-axis) in Fig.~\ref{fig:var}. Given a model trained on CIFAR-10 trained with 1000 labels and 49000 unlabeled data (denoted as $D_U$). We evaluate the accuracy on $D_U$ and test set $D_T$ respectively, and compute the five uncertainty metrics including PV, EV, PE, MI, and PR (defined in Section~\ref{subsec:metric}) by feeding forward test images 10 times under random dropout. Then, the input images are sorted in an ascending order according to their uncertainty values. Based on the five ranking lists, we distribute the input images of $D_U$ into 490 bins. In Fig.~\ref{fig:var} (a), each bin contains 100 images. The x-axis shows the bin index, and y-axis shows the average accuracy of the images that belong to the corresponding uncertainty ranked bin. Except predictive ratio~\cite{Gal2016Uncertainty}, other four metrics give similar trend of an inverse relationship between uncertainty and accuracy. Taking PV as an example, Fig.~\ref{fig:var} (b) show a strong inverse relationship between class accuracy and mean PV of each class on $D_T$. 

\begin{figure*}[t!]
	\centering
	\begin{subfigure}[t]{0.32\textwidth}
		\centering		
		\includegraphics[width=1\columnwidth]{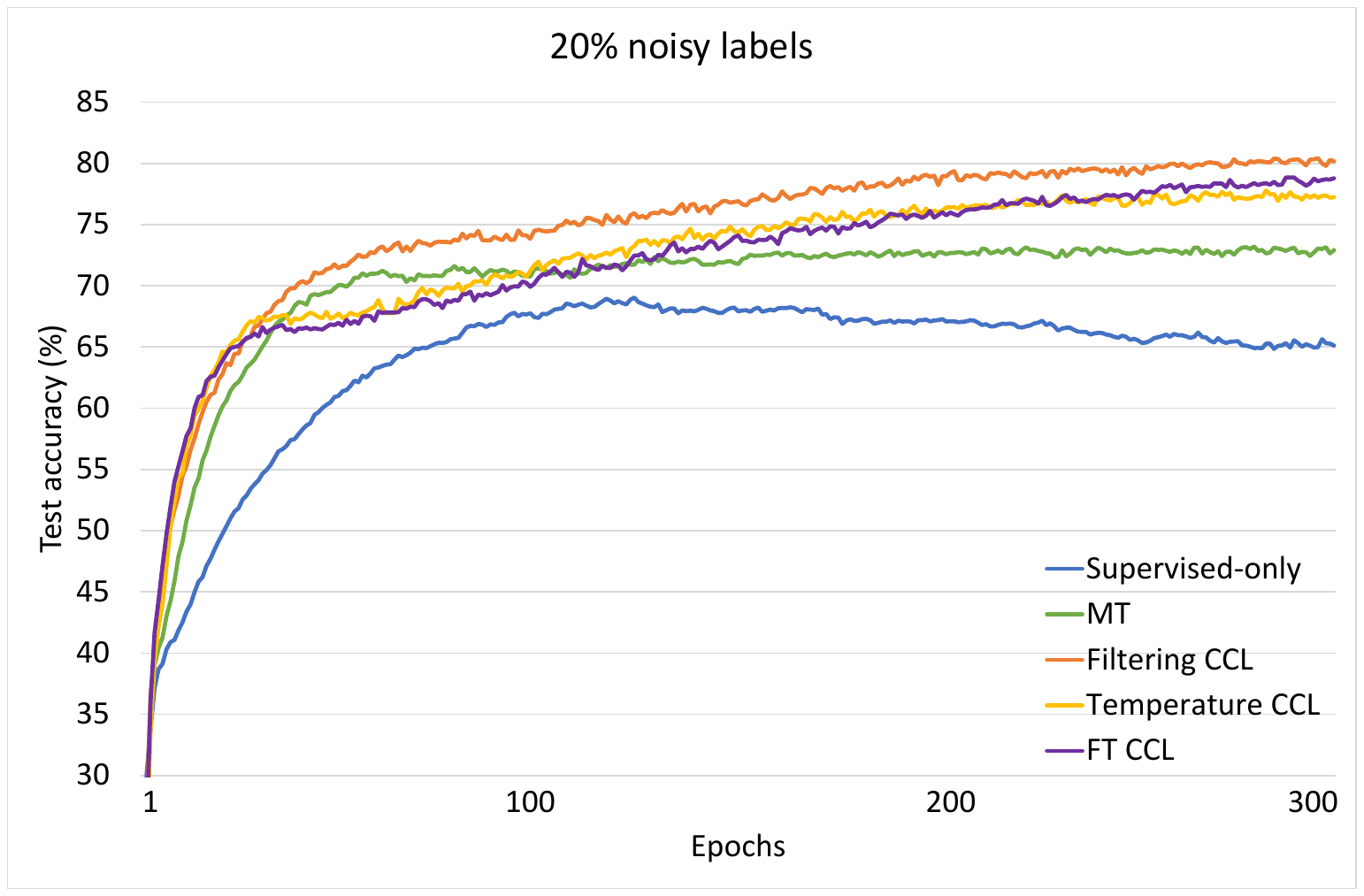}
	\end{subfigure}%
	~ 
	\begin{subfigure}[t]{0.32\textwidth}
		\centering
		\includegraphics[width=1\columnwidth]{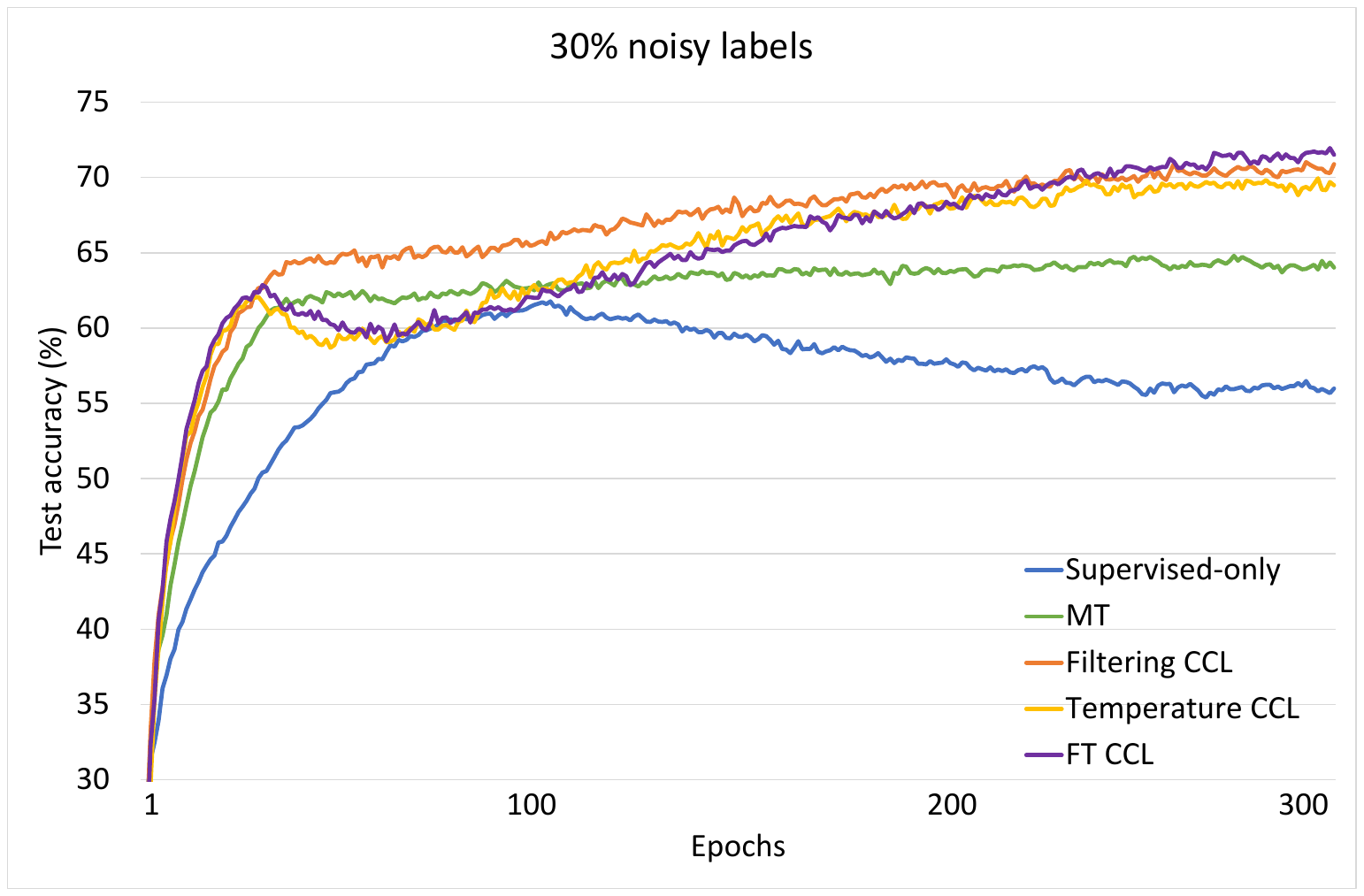}
	\end{subfigure} 
	~
	\begin{subfigure}[t]{0.32\textwidth}
		\centering
		\includegraphics[width=1\columnwidth]{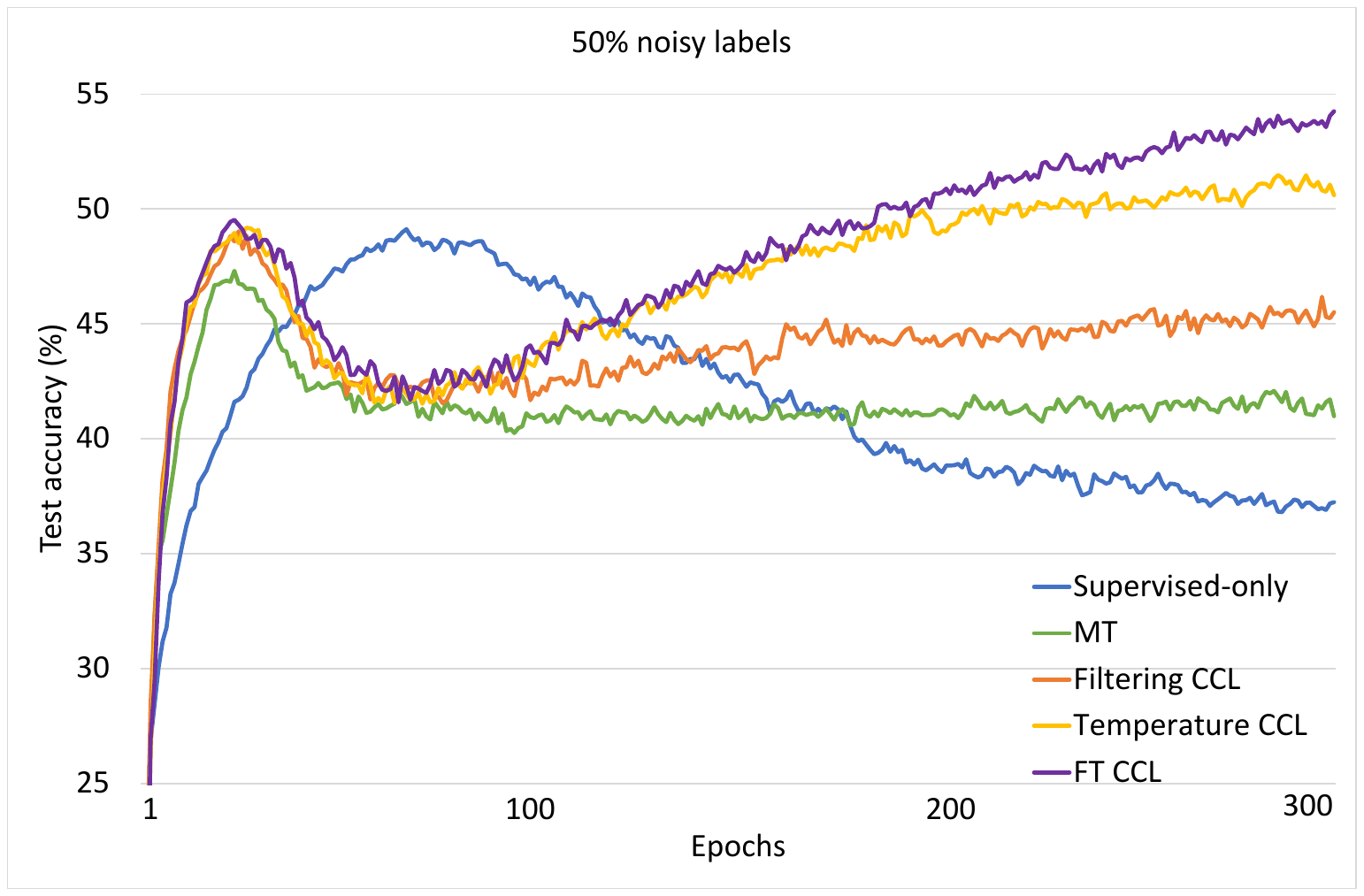}
	\end{subfigure}%
	\caption{Test accuracy on CIFAR-10 trained with 4000 labels with percentages of 20\%, 30\%, 50\% corrupted labels. Supervised only training (blue) fluctuates a lot and overfits to the incorrect labels. MT~\cite{tarvainen2017mean} shows the resistance to the corruption but still influenced by noisy labels. Our proposed Filtering CCL and Temperature CCL show considerable resistance to noisy labels. The high robustness of combined CCL verifies that our proposed methods provides complementary effects to prevent the model learning from uncertain/noisy targets. Better viewed in color.}
	\label{fig:noisy} 
\end{figure*}

\subsection{Robustness to Noisy Labels}

In a further test we studied the robustness of our method under random corruption of labels. Certain percentages (20\%, 30\%, 50\%) of true labels on the training set are replaced by random labels. Fig.~\ref{fig:noisy} shows the classification error on CIFAR-10 test set, trained with 4000 labels. In general, Filtering CCL performs better than Temperature CCL when noisy labels are few (e.g. 20\%). When the percentage of noisy labels increases, Temperature CCL outperforms Filtering CCL, and the combined FT-CCL shows the most robustness under 50\% noisy labels.

Specifically, when most of the labels are correct (20\% noisy), Filtering CCL is able to filter out uncertain noisy targets and gradually learn from the certain predictions. When there is moderate percentage (30\%) of labels are corrupted, Filtering CCL and Temperature CCL show comparable robustness. The former converges faster than the latter, and the FT-CCL outperforms both in the end. When 50\% of the labels are corrupted, Temperature CCL shows higher robustness than Filtering CCL, since uncertain targets remain high temperatures and low impacts in the consistency regularization compared to certain targets throughout training process. The FT-CCL further boosts the performance, indicating that our proposed Temperature CCL and Filtering CCL introduce complementary abilities to enforce consistency on reliable predictions. Thus, our proposed CCL provides considerable resistance to noisy labels, and improves the generalization performance of the model.

\section{Conclusion}
In this paper, we have proposed a certainty-driven consistency loss (CCL) to let the student learn meaningful and reliable targets from the teacher by utilizing the certainty information of the unlabeled data predictions. We propose two approaches Filtering CCL and Temperature CCL to filter out uncertain predictions in a hard way, and reduce the loss magnitudes of uncertain predictions in a soft way, respectively.
We conclude by stating that considering the predictive uncertainty of the unlabeled data is beneficial to semi-supervised classification, since the erroneous gradients coming from the uncertain predictions can be reduced. We further propose to decouple the student and teacher model to encourage model diversity, and train multiple student-teacher pairs in the network, which show further improvements. Nevertheless, as a trade-off, it requires to optimize multiple student models, and brings a larger memory footprint to store multiple student/teacher models. To the best of our knowledge, this is the first work that exploits the predictive uncertainty information into consistency regularization for semi-supervised learning. The extensive experiments on three semi-supervised benchmark datasets validate the effectiveness of our approach over a few existing methods. As a byproduct, our method can provide extra uncertainty information along with the predictions. Besides, our combined FT-CCL, that suppresses unreliable targets in both hard and soft way, show high tolerance to noisy labels, indicating that our proposed hard filtering and soft temperature adjusting strategies introduce complementary abilities to enforce consistency on reliable predictions. 
 
Interesting extensions of this work in the future may consist in: (1) Exploring stronger data perturbation techniques, e.g. adversarial examples~\cite{goodfellow2014explaining} and RandAugment~\cite{cubuk2020randaugment}. Stronger perturbations can  generate more and harder training samples, which can enable stronger consistency between student and teacher in the training process. 
(2) Utilizing uncertainty information into graph-based semi-supervised methods~\cite{luo2020every} to model graph-structured data. For example, one can modulate the label propagation procedure on the graph to let the signals coming from the uncertain nodes propagate less to the neighbouring nodes.
(3) Our idea of using uncertainty in the consistency loss can also be applied to other down-stream application tasks to control the contributes of unsupervised losses coming from different training samples. The down-stream tasks may include but not limited to semantic segmentation, domain adaptation, and person re-ID. Since the reliability of unlabeled predictions varies among different unlabeled data samples or image pixels, we believe that our proposed certainty-based loss modulation approach could provide useful recipe or idea for dealing with these tasks.


\bibliography{mybibfile}

\end{document}